# Gabor Filter Assisted Energy Efficient Fast Learning Convolutional Neural Networks

Syed Shakib Sarwar, Priyadarshini Panda, and Kaushik Roy
School of Electrical and Computer Engineering, Purdue University
West Lafayette, IN, USA
{sarwar,pandap,kaushik }@purdue.edu

*Abstract*— Convolutional Neural Networks (CNN) are being increasingly used in computer vision for a wide range of classification and recognition problems. However, training these large networks demands high computational time and energy requirements; hence, their energy-efficient implementation is of great interest. In this work, we reduce the training complexity of CNNs by replacing certain weight kernels of a CNN with Gabor filters. The convolutional layers use the Gabor filters as fixed weight kernels, which extracts intrinsic features, with regular trainable weight kernels. This combination creates a balanced system that gives better training performance in terms of energy and time, compared to the standalone CNN (without any Gabor kernels), in exchange for tolerable accuracy degradation. We show that the accuracy degradation can be mitigated by partially training the Gabor kernels, for a small fraction of the total training cycles. We evaluated the proposed approach on 4 benchmark applications. Simple tasks like face detection and character recognition (MNIST and TiCH), were implemented using LeNet architecture. While a more complex task of objet recognition (CIFAR10) was implemented on a state-of-the-art deep CNN (Network in Network) architecture. The proposed approach yields 1.31-1.53x improvement in training energy in comparison to conventional CNN implementation. We also obtain improvement up to 1.4x in training time, up to 2.23x in storage requirements, and up to 2.2x in memory access energy. The accuracy degradation suffered by the approximate implementations is within 0-3% of the baseline.

## I. INTRODUCTION

Deep learning Convolutional Neural Networks (CNNs) have proven to be very successful for various cognitive applications, notably in computer vision [1]. They have shown human like performance on a variety of recognition, classification and inference tasks, albeit at a much higher energy consumption. Few recent and well known CNN examples, deployed in real-world applications, are Google Image search, Google Now speech recognition [2], Microsoft's 'Project Adam' [3], Apple's Siri voice recognition and Google Street View [4]. However, the large scale structure and associated training complexity present CNNs as one of the most compute intensive workloads across all modern computing platforms. Hence, their energy-efficient implementation is of great interest.

In the recent past, approximate computing exploiting the intrinsic error resiliency of neural networks has been proposed to lower the compute effort [5]. A variety of approximate hardware and software techniques have been proposed to achieve computational efficiency [5,6]. However, most of the work aims at reducing the testing complexity of the network. In other words, the above techniques take a trained CNN and then impose approximations so that the compute effort can be scaled during testing. In contrast, the main focus of our work is to use error resiliency to reduce the training complexity of a CNN.

One of the major challenges for convolutional networks is the computational complexity and time needed to train large networks. Training of CNNs requires state-of-the-art accelerators like GPUs for extensive applications [7]. The large training overhead has restricted the usage of CNNs to clouds and servers. However, an emerging trend in IoT promises to bring the expertise of CNNs (image recognition and classification) to mobile devices that may lack or have only intermittent online connectivity. To ensure applicability of CNNs on mobile devices and widen the range of its applications, the training complexity must be reduced. Good amount of work has been done on speeding-up the training process through parallel processing [8], but not much work is found on improving the energy efficiency of CNN training. CNNs are trained using the standard back-propagation rule with slight modification to account for the convolutional operators [9]. The main power hungry steps of CNN training (back-propagation) are gradient computation and weight update of the convolutional and fully connected layers. In our proposed training, we achieve energy efficiency by eliminating a large portion of the gradient computation and weight update operations, with minimal loss of accuracy or output quality.

Classification accuracy is a primary concern for researchers in the machine-learning community. Different pre-processing models such as filters or feature detectors have been employed to improve the accuracy of CNNs. In fact, recent works employed Gabor filtering as a pre-processing step for training neural networks in pattern recognition applications [10,11]. Based on the human visual system, these filters are found to be remarkably appropriate for texture representation and discrimination. In [12, 13], the authors have attempted to get rid of the pre-processing overhead by introducing Gabor filters in the 1$^{st}$ convolutional layer of a CNN. In [12], Gabor filters replace the random filter kernels in the 1$^{st}$ convolutional layer. The training is then limited to the remaining layers of the CNN. The main focus of such work was accuracy improvement. However, these approaches may also lead to energy savings. In [13], the Gabor kernels in the 1$^{st}$ layer were fine-tuned with training. In other words, the authors used Gabor filters as a good starting point for training the classifiers, which helps in convergence. In this paper, we build upon the above works to propose a balanced CNN configuration, where fixed Gabor filters are not only in the 1$^{st}$ convolutional layer, but also in the latter layers of the deep CNN in conjunction with regular trainable convolutional kernels. Using Gabor filters as fixed kernels, we eliminate a significant fraction of the power-hungry components of the backpropagation training, thereby achieving considerable reduction in training energy. The inherent error resiliency of the networks allows us to employ proper blend of fixed Gabor kernels with trainable weight kernels to lower the compute effort while maintaining competitive output accuracy.

In summary, the key contributions of our work are as follows:

- We proposed the use of Gabor filters in different layers of the CNN to lower the computational complexity of training CNNs. The novelty of our work lies in the fact that we introduced Gabor filters with regular trainable weight kernels in the intermediate layers of the CNN.
- We developed an energy model for quantifying energy consumption of the network during training, based on the <u>M</u>ultiplication and <u>A</u>ccumulation (MAC) operations in the training algorithm.
- We designed several Gabor filter based CNN configurations in order to get the best trade-off between accuracy and other parameters of interest, especially energy consumption.
- We mitigate accuracy loss of Gabor filter based CNNs by partially training the Gabor kernels.

We show that our proposed methodology leads to energy efficiency, reduction in storage requirements and training time, with minimal degradation of classification accuracy.

## II. CONVOLUTIONAL NEURAL NETWORK: BASICS

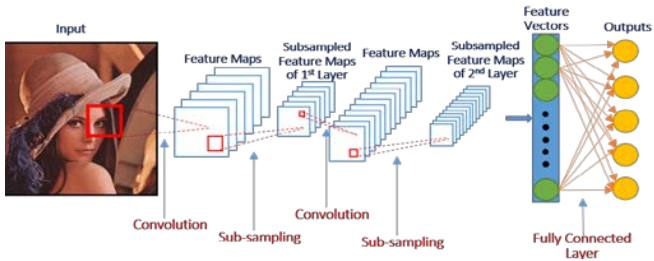

Fig. 1. A standard architecture of a deep learning convolutional neural network.

Convolutional neural networks consist of convolutional and fully connected layers, with nonlinearity applied at the end of each layer. They may also include pooling layers, which combine the outputs of neuron clusters. To improve generalization and to reduce the number of trainable parameters, a convolution operation is exercised on small regions of input. One salient benefit of CNNs is the use of shared weights in convolutional layers, implying that the same filter (weight bank) is used for each pixel of the image; this reduces memory footprint and enhances performance. A standard architecture of a deep learning CNN is shown in Fig. 1.

The basic operation of CNNs consists of two stages: i) Training and ii) Testing. The training process is usually carried out using back-propagation algorithm [9]. The trained CNN is then used to test random data inputs. The testing process is basically forward propagation, which is much simpler compared to training, from energy consumption and complexity aspect. For deep networks with huge training datasets, the training process may require significant computation power. As mentioned earlier, the most power consuming steps of a CNN training with back-propagation are, gradient computation, and weight update of the convolutional and fully connected layers, which by far outweighs other steps such as error and loss function computation. In this work, we provide a solution to achieve energy efficiency in training by getting rid of a significant portion of the power hungry gradient computation and weight update operations, with minimal loss of accuracy.

## III. GABOR FILTERS

Many pattern analysis applications such as character recognition, object recognition, and tracking require spatially localized features for segmentation. Gabor filters are a popular tool for extracting these spatially localized spectral features [14]. Simple cells in the visual cortex of mammalian brains can be modeled by Gabor functions [15]. Frequency and orientation representations of Gabor filters are similar to those of the human visual system, and they have been found to be particularly appropriate for texture representation and discrimination. A particular advantage of Gabor filters is their degree of invariance to scale, rotation, and translation. In fact, [16] have shown that deep neural networks trained on images tend to learn first layer features resembling Gabor filters. This further corroborates our intuition of using pre-designed Gabor filters as weight kernels in a CNN configuration.

An appropriately designed Gabor filter extracts useful features corresponding to an input image. However, this would also require us to have separate uniquely designed filters for every image that will be infeasible for large-scale problems. In order to have a generic approach, we have used a systematic method where filters are generated using a 'filter bank' [14, 17]. The generated Gabor filters are then used to replace the weight kernels in the CNN configuration. We have used real values of 2D Gabor filters as weight kernels. Also, the filters used in our methodology are equally spaced in orientation (for instance, θ=0°, 30°, 60°, 90° etc.) to capture maximum number of characteristic textural features.

## IV. DESIGN APPROACH & METHODOLOGY

The use of fixed Gabor kernels to exploit error resiliency of CNN, is the main concept of our proposed scheme. This section outlines the key steps of the proposed design methodology.

### A. Energy Model for CNN Training

To realize the effect of Gabor kernels on energy, we developed an energy model to get the distribution of energy consumption for training the network. The energy model is based on the number of Multiply and Accumulate (MAC) operations in the training algorithm. The MAC circuits were designed in Verilog and mapped to 45 nm technology using Synopsys Design Compiler. Then the power and delay numbers from the Design Compiler were fed to the energy computation model to determine the distribution of energy consumption.

From the energy distribution for training a CNN, we found that in a conventional CNN containing 2 convolutional layers each followed by a sub-sampling layer, and finally a fully connected layer ([784 6c 2s 12c 2s 10o]), the 2$^{nd}$ convolutional layer uses 27% of the overall energy consumption during training, while the 1$^{st}$ convolutional layer consumes 20%. We realized that, in order to achieve energy efficiency, the energy spent on these layers needs to be minimized, and using Gabor filters as fixed kernels is the key.

### B. Gabor Filters in First Convolutional Layer

We designed a CNN, by replacing certain weight kernels with fixed Gabor filters. The network has 2 convolutional layers each followed by a sub-sampling layer, and finally a fully connected layer. The 1$^{st}$ convolutional layer has $k$ kernels, while the 2$^{nd}$ convolutional layer has $2k$ kernels for each of the $k$ feature maps of the 1$^{st}$ layer, resulting in total of $2k^2$ kernels in the 2$^{nd}$ layer. In this specific example, we consider $k$=6. Hence, for our example, the 1$^{st}$ convolutional layer consists of 6 kernels and the 2$^{nd}$ convolutional layer consists of 72 kernels. The 12 feature maps from the 2$^{nd}$ layer are used as feature vector inputs to the fully connected layer which produces the final classification result. We used 6 Gabor kernels (5×5 sized), which are equally spaced in orientation, (with θ=0°, 30°, 60°, 90°, 120° and 150°) to replace the regular kernels of the 1st convolutional layer. The network was trained on MNIST dataset [18]. The regular trainable kernels of the 1st layer of the CNN after 100 epochs and fixed Gabor kernels are shown in Fig. 2 to have a comparative view. The results of this configuration are shown in Table I (Row 2 corresponding to Fixed Gabor/Trainable CNN configuration).

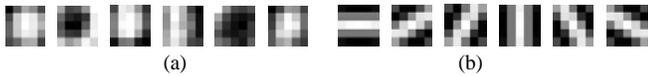

(a)          (b)

Fig. 2. (a) Trained kernels in 1st convolutional layer. (b) Fixed Gabor kernels equally spaced in orientation.

From Row 2 in Table I, it can be clearly observed that the Fixed Gabor/Trainable CNN configuration has an accuracy comparable to the conventional CNN implementation with a marginal loss of 0.62% (baseline accuracy 99.09%). In addition, we observe 20.7% reduction in energy consumption, and 9.47% decrease in training time. The storage requirements for the network's trainable parameters remain unchanged.

### C. Gabor Filters in Both Convolutional Layers

To achieve higher energy improvements, we turned the convolutional layers all Gabor, i.e. all the kernels in both the convolutional layers of the CNN were replaced with fixed Gabor filters. In the 2nd convolutional layer, there are 12 kernels for each of the 6 output feature maps of the 1st convolutional layer, in total 72 kernels. We used the same 6 Gabor kernels from section IV.A, equally spaced in orientation, in the 1st convolutional layer, and 12 Gabor kernels, equally spaced in orientation (with θ=0°, 15°, 30°, 45°, 60°, 75°, 90°, 105°, 120°, 135°, 150° and 165°), in the 2nd convolutional layer. Considering the Gabor kernels as constants, 12 Gabor kernels are sufficient to replace the 72 kernels (convolved output of which are summed to produce 12 feature maps) of the 2nd layer. In this case, same set of 12 Gabor kernels will convolve with each of the 6 output feature maps of 1st layer in every training cycle. It is evident that the 6 kernels of the 1st layer are also present in the 2nd layer. This actually helps to carry the integrity of the previous layer to the following layer. The benefits observed from this Fixed Gabor/Fixed Gabor CNN configuration can be seen from Row 3 in Table I. The training time reduced by 53%, leading to 1.93x improvement in energy consumption; also the storage requirement is drastically reduced. However, the big downside is the high accuracy degradation (5.85%), even for a simple dataset like MNIST. The configuration leads to intolerable accuracy degradation for more complex datasets.

TABLE I. COMPARISON BETWEEN DIFFERENT CNN CONFIGURATIONS

| Configuration | | Accuracy Loss | Energy Savings | Training Time Reduction | Storage Savings |
|---|---|---|---|---|---|
| 1st Conv. Layer Kernels | 2nd Conv. Layer Kernels | | | | |
| Trainable | Trainable | -- | -- | -- | -- |
| Fixed Gabor | Trainable | 0.62% | 20.70% | 9.47% | 0% |
| Fixed Gabor | Fixed Gabor | 5.85% | 48.28% | 53% | 42.48% |
| Fixed Gabor | Half Fixed Gabor & Half Trainable | 1.14% | 34.49% | 22.30% | 23.15% |

*Accuracy loss, energy savings, training time reduction and storage savings are computed by considering the conventional CNN results as baseline. The baseline accuracy is 99.09%.

### D. Balanced Network Configuration for Maximum Benefits

As seen in section IV.A, not training the 1st layer kernels incurs minor accuracy loss. However, from section IV.B we see that, fixing the 2nd layer reduces the accuracy drastically, while it gives very high energy savings. In order to avoid such accuracy degradation, we designed a blended CNN configuration where the 2nd convolutional layer is partly trained with a combination of fixed Gabor filter kernels and regular trainable weight kernels, while the 1st convolutional layer uses fixed Gabor kernels only. However, there can be multiple CNN configurations for different number of fixed and trainable kernels in the 2nd layer, each providing different amount of energy savings for corresponding accuracy degradation.

To obtain the optimal configuration, we made an effort to balance the trade-off between accuracy and other parameters of interest, especially energy consumption. We conducted an experiment where we trained 5 configurations of blended CNN on MNIST dataset with $i$ number of fixed Gabor kernels in the 2nd layer where $i$= 0, 3, 6, 9, 12. Each of the configurations was trained for 100 epochs. Fig. 3 shows the overall classification accuracy obtained with the blended CNN configuration as we varied the number of trainable weight filters in the 2nd layer. Again, as in section IV.A, Gabor kernels with equally spaced in orientation were used. It can be clearly observed that the accuracy degrades with the increase in the number of fixed Gabor filters. To select the optimal configuration, we imposed a constraint of maximum 1% degradation in the classification accuracy. In the plot (Fig. 3), the solid square corresponds to the accuracy for the Fixed Gabor/Trainable CNN configuration (Row 2 of Table I) and the solid circle is the point where degradation in classification accuracy is ~1%. This solid circle corresponds to a half-half configuration where half of the kernels in the 2nd layer is fixed while the other half is trainable. Fig. 3 also shows the energy savings, training time reduction and storage requirement savings observed with different configurations. As expected, the benefits (energy, training time, storage) increase as we fix more filters, beyond the half-half point. However, the decrease in accuracy is significant. We observed similar trend for CNNs trained on FaceDet and TiCH [19] datasets as well for this LeNet architecture. Based on this, we used the half-half configuration to implement the CNN for MNIST, FaceDet and TiCH, in order to get maximum benefits with minimal degradation in accuracy. The results are described in the following section.

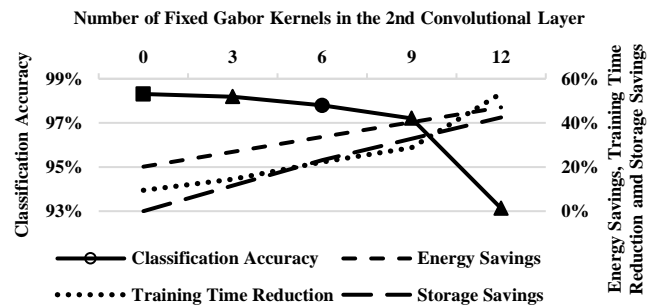

Fig. 3. Change in classification accuracy, energy savings, training time reduction and storage requirement with different configurations of fixed Gabor kernels in the 2nd convolutional layer.

The half-half balanced configuration implies that we have 6 fixed and 6 trainable kernels (in the 2nd layer) for each of the 6 feature maps from the 1st layer. Thus, total trainable kernels are 36 (6×6). Since the other 36 (6×6) kernels, for each of the 6 feature maps from the 1st layer, are fixed, they can be replaced by 6 Gabor kernels instead of 36. Again, the 6 fixed Gabor kernels used for the 2nd layer are same as the 6 Gabor kernels of 1st layer. Using the same kernels across the 1st and 2nd layer has two-fold advantage: i) it helps to carry the integrity from the previous layer to the following layer. This means, the network can learn new features in the 2nd layer on top of the features it learned in the 1st layer, ii) the storage requirement is highly optimized. We do not need any new Gabor kernel for the replacement of 36 weight kernels of the 2nd layer. Using different Gabor kernels in the two layers gives slightly lesser accuracy with a slightly higher storage requirement.

Fig. 4 gives a comparative picture of the regular trainable kernels of the conventional CNN and half fixed/half trainable kernels of the proposed balanced system, in the 2nd layer after 100 epochs of training. The results of the Half Fixed/Half Trainable configuration are shown in Row 4 of Table I. It is evident that the accuracy loss is tolerable (1.14%). Also, we see a significant reduction in energy, training time and storage requirements.

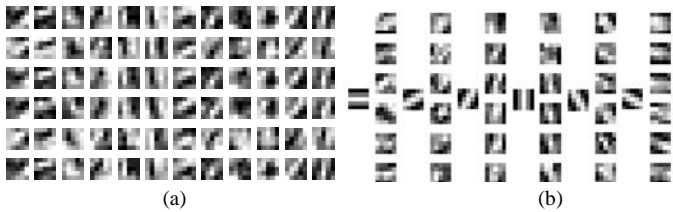

Fig. 4. (a) Trained kernels in 2nd convolutional layer and (b) Kernels of the proposed half fixed/half trainable configuration in 2nd convolutional layer.

From this analysis, we observe that Half Fixed/Half Trainable configuration is optimal for the network of this structure. However, other networks with different structures will have different optimal configurations. Various combination of fixed-trainable Gabor kernels can be used as a tuning knob in deeper networks to meet the energy-quality requirements.

## V. SIMULATION FRAMEWORK

This section describes the overall simulation framework. We used modified versions of open source MATLAB codes [20,21] to implement multilayer CNNs for our experiments. We trained the CNNs using the corresponding training datasets mentioned in Table II, to get the accuracy and training time information, which were used as a baseline for comparison. Then we introduced Gabor filters as weight kernels, and modifications in the training algorithm were made, so that the advantage of Gabor kernels can be realized. We used these newly formed CNNs for training with the corresponding datasets and collected the accuracy and training time information.

We developed an energy computation model based on the number of MAC operations in the training algorithm. We implemented multiplier and adder units at the Register-Transfer Level (RTL) in Verilog and mapped them to the IBM 45nm technology in 1 GHz clock frequency using Synopsys Design Compiler. The power and delay numbers from the Design Compiler were fed to the energy computation model to get energy consumption statistics. We also computed storage requirements and memory access energy for the overall network based on input size, number of convolutional layers, number of kernels in each layer, size of fully connected layer, number of neurons in the fully connected layer and number of output neurons. Details of the benchmarks used in our experiments are listed in Table II:

TABLE II. BENCHMARKS

| Application | Dataset | Number of Training Samples | Number of Testing Samples | Input Image Size |
|---|---|---|---|---|
| 1. Face Detection | Face-Nonface | 600 | 200 | 48×48 |
| 2. Digit Recognition | MNIST | 60000 | 10000 | 28×28 |
| 3. Tilburg Character Set Recog. | TICH | 30000 | 10000 | 28×28 |
| 4. Object Recognition | CIFAR10 | 50000 | 10000 | 32×32 |

CNN architectures used for the experiments are listed in Table III:

TABLE III. CNN ARCHITECTURES

| Dataset | Architecture | Network |
|---|---|---|
| Face-Nonface | LeNet [22] | [784 (5×5)6c 2s (5×5)12c 2s 2o] |
| MNIST |  | [784 (5×5)6c 2s (5×5)12c 2s 10o] |
| TICH |  | [784 (5×5)10c 2s (5×5)20c 2s 36o] |
| CIFAR10 [23] | NIN [24] | [1024×3 (5×5)192c 160fc 96fc (3×3)mp (5×5)192c 192fc 192fc (3×3)mp (3×3)192c 192fc 10o] |

*c: convolutional layer kernels, s: sub-sampling kernel, fc: fully connected layer neurons, mp: max pooling layer kernel, o: output neurons

## VI. RESULTS

In this section, we present the benefits of our proposed design (with half-half balanced configuration for FaceDet., MNIST and TiCH). The results of the conventional CNN implementation, in which all of the convolutional kernels are trainable, are considered as baseline for all comparisons. All the comparisons are done under iso-epoch condition.

### A. Accuracy Comparison

Fig. 5 shows the classification accuracy obtained after 100 epochs, using conventional CNN and proposed Gabor kernel based CNN for various applications.

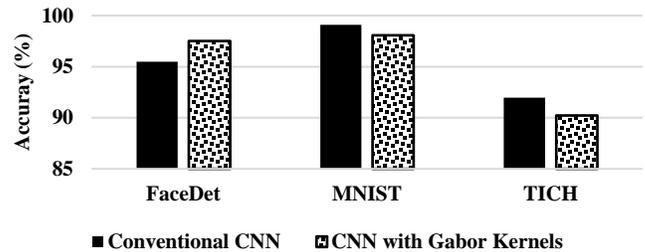

Fig. 5. Comparison of accuracy between conventional CNN and Gabor kernel based CNN for different applications.

An interesting thing to note is that for FaceDet, the accuracy of the proposed CNN is better than the baseline. This can be attributed to the fact that FaceDet is a simpler detection task where we need to detect faces from a collection of face and non-face images in the dataset. Gabor filters being edge detectors, further simplify the face detection problem [11]. Thus, we observe better accuracy on FaceDet. In contrast, other benchmarks are multi-object classification problems where we need to predict the class label from a collection of objects. Among the remaining benchmarks, MNIST and TiCH are character recognition datasets. The accuracy baselines for the datasets FaceDet., MNIST and TiCH are 95.5, 99.09 and 91.98 respectively. The accuracies reported in this work are obtained using test datasets, which are separate from the training datasets for each of the benchmarks.

## B. Energy Consumption Benefits

Fig. 6 shows the improvement in energy consumption achieved (during training), using Gabor kernel based CNN for different applications. We achieve 31-35% energy savings for training of CNNs for first three benchmarks mentioned in Table II.

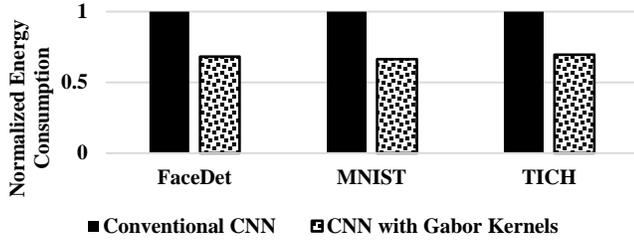

Fig. 6. Comparison of energy consumption during training, between conventional CNN and Gabor kernel based CNN for different applications.

In Fig. 7, the pie chart represents a sample energy distribution during training a CNN, across different segments. The CNN used for the analysis contains two convolutional layers and two fully connected layers, and was trained on MNIST dataset. It can be seen that computation of error, loss function [9], and back propagation of errors consume small fraction (<1%) of the total energy. A sizeable portion of the energy consumption is captured by the forward propagation, gradient computation and weight update at the convolutional and fully connected layers. During back-propagation, for the $1^{st}$ convolutional layer this energy consumption is 20% of the total energy consumption, while for the $2^{nd}$ convolutional layer it is 27%. Considering that the $1^{st}$ convolutional layer contains only fixed Gabor kernels, the entire 20% energy consumption, required for $1^{st}$ convolutional layer during back-propagation (as mentioned earlier), can be saved. Also in the $2^{nd}$ convolutional layer, half of the kernels is trainable. That means for the $2^{nd}$ convolutional layer, we need 13.5% of the training energy compared to the 27% energy requirement of a conventional CNN implementation. Therefore, we can save up to 33.5% of the training energy in this particular example. This is due to the fact that we do not have to perform gradient computation or weight update for the fixed Gabor weight kernels.

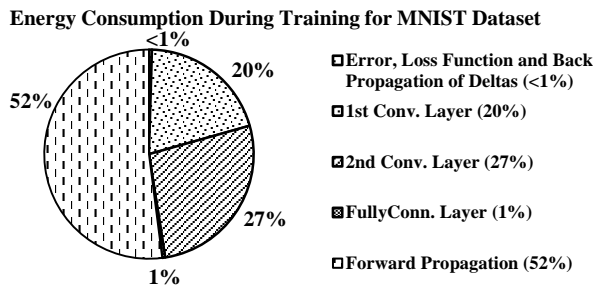

Fig. 7. Energy consumption of different segments during training of a CNN with MNIST dataset.

The amount of energy benefit is dependent on the network structure. If the fully connected layer is much larger than the convolutional layers with many hidden layers, then the energy savings will be less. Again, if we have more convolutional layers with fixed Gabor kernels, we will get larger energy savings. In case of a deep CNN with more than two convolutional layers, the selection of the number of fixed kernels in different layers will depend on several key factors such as network structure and size, complexity of the dataset, quality requirements, among others.

## C. Storage Requirement Reduction

Fig. 8 shows the storage requirement reduction obtained using proposed scheme for different applications. We achieve 15-30% reduction of storage requirement across the various benchmarks (Table II).

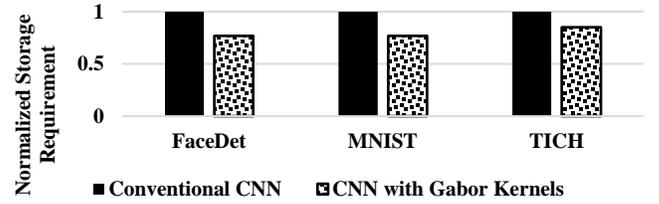

Fig. 8. Comparison of storage requirements between conventional CNN and Gabor kernel based CNN for different applications.

A large part of the training energy is spent on the memory read/write operations. Proposed Gabor filter assisted training also provides savings in 1.2-1.3x memory access energy since we do not need to write (update during backpropagation) the fixed kernel weights during training.

## D. Training Time Reduction

Since gradient computation and weight update is not required for the fixed Gabor kernels, we achieve significant savings in computation time with our proposed scheme. Fig. 9 shows the normalized training time per epoch for each application. We observe 10-29% reduction in training time per epoch across the first three benchmarks of Table II.

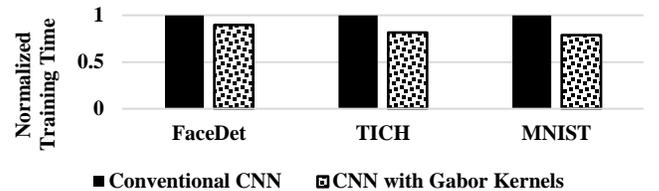

Fig. 9. Comparison of training time requirements between conventional CNN and Gabor kernel based CNN for different applications.

Point to be noted that the training time reduction is a by-product of our proposed method. This spare time cannot be used to recover accuracy loss by providing more epochs to the training as it will absorb the energy savings.

## E. Partial Training of Gabor Kernels for Accuracy Improvement

Gabor kernel based CNN provides energy savings at a cost of some accuracy degradation. For complex recognition applications, the accuracy degradation is higher. Such problems can be mitigated by employing partial training of the Gabor kernels. In the partial training, we start the CNN training with the Gabor kernels in respective positions, and then allow them to learn for first few iterations (say 20% of total number of training cycles). Since Gabor kernels provide a good starting point for the training, the kernels will learn quickly and reach near saturation within a small percentage of training cycles. Then we stop updating the partially learned Gabor kernels for the remainder of the training cycles. Point to be noted here that partial training introduces overhead, slightly increasing the energy consumption, the average training time, the memory access energy, and the storage requirement compared to the Fixed Gabor kernel based CNN implementation. Nevertheless, partial training provides substantial energy saving over conventional CNN implementation, and can be a used as another tuning knob for the deep CNNs to trade-off energy-quality.

*F. Applicability in Complex CNNs*

Employing fixed Gabor kernels is also beneficial for larger and more complex CNNs containing more than two convolutional layers. To corroborate that, we implemented a deep CNN: [1024x3 (5x5)192c 160fc (3x3)mp (5x5)192c 192fc 192fc (3x3)mp (3x3)192c 192fc 10o] (c: convolutional layer kernels, fc: fully connected layer neurons, mp: max pooling layer kernel) for CIFAR10 [23] dataset using MatConvNet [21]. The network contained three MlpConv[24] blocks, each of which contained one convolutional layer, two fully connected layers and one max-pooling layer. Using 128 (2/3$^{rd}$ of the total 192) fixed Gabor filters as convolutional kernels in each of the convolutional layers, we achieved ~29% energy savings and 55.7% storage reduction, while losing ~3.5% classification accuracy compared to conventional CNN implementation. Though the energy savings is impressive, nevertheless the accuracy degradation is higher than desired. Therefore, we employed partial training to salvage fraction of the dropped accuracy. The Gabor kernels of the 1$^{st}$ MlpConv block were kept fixed. The Gabor kernels of the 2$^{nd}$ and 3$^{rd}$ MlpConv block were partially trained for (20-30%) of total number of training cycles. This led to accuracy improvement of ~1.5%, while the overhead of partial training was not substantial. The results of such training are listed in Table IV.

TABLE IV. COMPARISON BETWEEN DIFFERENT TRAINING CONFIGURATIONS

| Configuration | | Percentage of total Training Cycles | Accuracy Loss | Comp. Energy Savings | Storage Requirement Savings | Memory Access Energy Savings |
|---|---|---|---|---|---|---|
| 2$^{nd}$ Block Gabor Kernels | 3$^{rd}$ Block Gabor Kernels | | | | | |
| Fixed | Fixed | 0% | 3.34% | 29.1% | 55.2% | 54.97% |
| Fixed | Trained | 20% | 2.5% | 28.94% | 32.31% | 50.55% |
| Trained | Trained | 20% | 2.41% | 25.7% | 0.5% | 44.19% |
| Trained | Trained | 30% | 1.95% | 24.04% | 0.5% | 38.71% |

*Accuracy loss, energy savings, storage savings and memory access energy savings are computed by considering the conventional CNN results as baseline. The baseline accuracy is 88.5%.

The proposed scheme is scalable for deeper networks and can be employed in variety of classification tasks. The energy-accuracy trade-off heavily depends on the network structure, the complexity of the task, the sensitivity of the convolutional layers, the number of fixed kernels in different layers, among others.

## VII. CONCLUSION

In recent times, deep learning methods have outperformed traditional machine learning approaches on virtually every single metric. CNNs are one of the chief contributors to this success. To meet the ever-growing demand of solving more challenging tasks, the deep learning networks are becoming larger and larger. However, training of these large networks require high computational effort and energy requirements. In this work, we exploited the error resiliency of CNN applications and the usefulness of Gabor filters to propose an energy efficient and fast training methodology for CNNs. We designed and implemented several Gabor filter based CNN configurations to obtain the best trade-off between accuracy and energy. We proposed a balanced CNN configuration, where fixed Gabor filters are not only used in the 1$^{st}$ convolutional layer, but also in the latter convolutional layers in conjunction with regular weight kernels. Experiments across various benchmark applications with our proposed scheme demonstrated significant improvements in energy consumption during training, and also reduction in training time, storage requirements, and memory access energy for negligible loss in the classification accuracy. Note, since our proposed Gabor kernel based CNN is faster and consumes less energy during training, the cost of retraining the network, when new training data is available, will be less compared to a conventional CNN. Also since fixed kernels are used in 50-66.7% convolution operations, dedicated hardware can be designed to gain more benefits not only in back-propagation, but also during forward-propagation in the network.